\documentclass{article}

\newcommand{\add}[1]{\color{black}{{#1}}} 
\newcommand{\rem}[1]{\ignorespaces}

\pdfoutput=1
\usepackage{arxiv}

\usepackage[utf8]{inputenc} 
\usepackage[T1]{fontenc}    
\usepackage{hyperref}       
\usepackage{url}            
\usepackage{booktabs}       
\usepackage{amsfonts}       
\usepackage{nicefrac}       
\usepackage{microtype}      
\usepackage{lipsum}
\usepackage{graphicx}
\usepackage{xcolor}

\title{Incremental Real-Time Personalization in Human Activity Recognition Using Domain Adaptive Batch Normalization}

\author{
  Alan Mazankiewicz \\
  Karlsruhe Institute of Technology\\
   \And
  Klemens Böhm \\
  Karlsruhe Institute of Technology \\
   \And
  Mario Bergés \\
  Carnegie-Mellon-University
}

\begin{document}
\maketitle

\begin{abstract}
    Human Activity Recognition (HAR) from devices like smartphone accelerometers is a fundamental problem in ubiquitous computing.
    Machine learning based recognition models often perform poorly when applied to new users that were not part of the training data.  
    Previous work has addressed this challenge by personalizing general recognition models to the unique motion pattern of a new user in a static batch setting. 
    They require target user data to be available upfront.
    The more challenging online setting has received less attention.
    No samples from the target user are available in advance, but they arrive sequentially.
    Additionally, the motion pattern of users may change over time.
    Thus, adapting to new and forgetting old information must be traded off.
    Finally, the target user should not have to do any work to use the recognition system by, say, labeling any activities. 
    Our work addresses all of these challenges by proposing an unsupervised online domain adaptation algorithm. 
    Both classification and personalization happen continuously and incrementally in real time.
    \add{Our solution works by aligning the feature distributions of all subjects, be they sources or the target, in hidden neural network layers.
    To this end, we normalize the input of a layer with user-specific mean and variance statistics. 
    During training, these statistics are computed over user-specific batches. 
    In the online phase, they are estimated incrementally for any new target user.}
    \rem{Our solution works by aligning the feature distribution of all the subjects, sources and target, within deep neural network layers.}  
    \rem{Experiments with 44 subjects show accuracy improvements of up to 14 \% for some individuals.
    Median improvement is 4 \%.}
\end{abstract}

\keywords{human activity recognition, convolutional neural networks, batch normalization, transfer learning, online learning, incremental personalization, online domain adaptation}

\section{Introduction} \label{sec:introduction}
Human Activity Recognition (HAR) is a fundamental building block for many emerging services such as health monitoring, smart personal assistance or fitness tracking.
These activities are often detected with mobile wearable sensors, like accelerometers in smartphones, and classified by a pre-trained machine learning model \cite{lockhart2012applications}. 
A challenge is to personalize a general recognition model trained on a set of source users to a new unseen target user. 
Without any personalization, general models often perform poorly on unseen target users \cite{weiss2012impact}, \cite{jordao2018human}.
This is because many users have a unique motion pattern that leads to a significant shift between the distributions of the sources and the target.
\add{Machine learning models however assume that the distribution during training and application (testing) stays the same.}
Much previous work has addressed this challenge in a static batch setting, i.e., the full target user data is available as a batch at once.
This current work addresses the personalization problem in the more challenging online setting.
Specifically, we assume that the samples from a new, unseen target user arrive sequentially, possibly until infinity. 
Here, classification and personalization need to happen continuously in real time (e.g., every 1-2 seconds) based on few current observations.
An algorithm also cannot store all previous observations and retrain the model each time with the entire batch. 
Instead, it must update its state incrementally based on the new observations and possibly disregard them afterwards. 
Furthermore, the motion pattern of the target user for an activity or the operation setting of the system does not necessarily remain the same over time.
Thus, the algorithm must ``unlearn'' old information and adapt to new one \cite{gepperth2016incremental}. 
\add{The online setting is particularly important for HAR in systems that are context-aware.
There, a system needs to accurately detect what a user is doing in real time so that it can react to his actions accordingly.}
To tackle these problems, some previous work has proposed solutions based on active learning \cite{settles2010active}, variants of self-learning \cite{triguero2015self} or a combination of the two. 
\add{We see the following drawbacks that our work explicitly addresses:}

Solutions based on active learning try to identify a few target instances where ground truth information would increase classification accuracy significantly.
In a real application, the label would be obtained by asking the user, which could decrease the overall satisfaction and willingness to use the recognition system.
\add{Further, a mistake can decrease the detection accuracy of the model.}
Self-learning based solutions on the other hand classify the target data and update the model by assuming that very confident predictions are indeed true.
While the user does not have to label any data, motion patterns between users can be very different. 
As self-learning is not a transfer learning algorithm designed to adapt to such significant distribution shifts, it is likely to yield erroneous results under these circumstances. 
Another shortcoming of previous work is that personalization does not happen simultaneously with classification in real time. 
Instead, it is postponed until a larger batch is collected.
We expect that using newly available target user data right away may improve classification earlier. 
Finally, previous work does not address the stability-plasticity dilemma that occurs in the online setting i.e., regulating the trade-off between forgetting old information and adapting to new one \cite{gepperth2016incremental}.
Depending on the setting, adaptation must be stronger or weaker. 

\add{Our solution addresses all of the issues just mentioned.
Our goal is to align the feature distribution of all users so that a model trained on the sources can work well on the target. }
\rem{Our solution addresses these issues.}
\add{As the  general model trained on data from the source users
we use a neural network.} 
\rem{We use a convolutional neural network (CNN) as the initial general model trained on a set of source users to classify the incoming sensor stream of a target user.}
The network contains an incremental online extension of adaptive batch normalization for domain adaptation (DA-BN) \cite{mancini2018kitting, li2016revisiting} \add{in its fully connected layers}.
During training, the only difference to a standard neural network with \add{fully connected} batch normalization (BN) layers \cite{ioffe2015batch} is that \rem{each batch only consists of data from the same person} \add{there is only data from the same person in a given batch. 
Therefore, data from different source users is in different batches.}
\add{Given this, each batch is normalized using a user-specific mean and variance estimate (BN statistics). 
We assume that during training the distribution of sensory measurements of a user stays the same, while it differs between users.
Normalizing the batches of each user within the network layers with the individual statistics should align the feature distributions of all users.
So the distributions of the output of each fully connected layer should be the same independent of the user. 
Figure \ref{fig:cross_da} displays the intuition.}
\rem{Given this, each batch is normalized by user-specific BN statistics.}
In the online testing phase, sliding windows \add{containing accelorometer measurements} from a new unseen target user arrive and are classified in real time. 
Since the target user has never been seen by the model before, the user-specific statistics for the normalization are unknown. 
\rem{Initially, these statistics are estimated from the source users during training.}
\add{They are initialized as the average over all sources.}
\rem{In the online phase} \add{As target user data sequentially arrives}, these initial values are updated incrementally from each single sliding window, right before classifying the respective window. 
Thus, the statistics gradually adjust to the target user with each additional instance.
This makes the feature distribution of the target overlap with the sources from training. 
\cite{mancini2018kitting} were first to propose an online version of DA-BN. 
We in turn use an incremental exponential variance formulation proposed by \cite{finch2009incremental} that updates the global statistics based on a single instance, instead of a batch. 

We summarize the advantages of our method as follows: 
\begin{itemize}
    \item Personalization happens in real-time, i.e., no batch of different activities from the target person has to be collected before or during the online phase. 
    Instead, an incremental personalization step happens each time right before classifying an instance. 
    Further, the user is never asked to label any data before or during the online phase. 
    \item Our method processes a (potentially) infinite sequence of measurements with constant memory and performs incremental updates efficiently with one pass over the data. 
    \item Our method is adaptive to changes in user motion or the operation setting of the system over time (concept drift). 
    The exponential average and variance formulation provides a parameter $\alpha$ to regulate how gradual adaptation to such changes should be. 
    \item Our method models the personalization problem as a theoretically grounded online unsupervised transfer learning problem.
    This allows it to deal with distribution shifts between source and target by design. 
    \item Our method is the first deep-learning based approach applied to personalization in online HAR. 
    Deep learning based models have shown to outperform shallow models in HAR and many other machine learning tasks.
\end{itemize}{}

In our experiments with 44 subjects on 5 activities, we observe improvements in accuracy over our baseline \add{without personalization} of up to 14 \%  for some individuals. Median improvement \add{over all subjects} is 4 \%.

\section{Related Work} \label{sec:related_work}
In this section, we first cover personalization for HAR in general, before reviewing current online personalization approaches.

\cite{twomey2018comprehensive} is a general introduction to HAR. 
In HAR, general models usually are trained from a set of source users for whom labeled measurements have been collected for training.
Personalization aims at adjusting the general model to accurately classify the activities of a new unseen target user whose motion patterns may be quite different from the source users.   

\cite{weiss2012impact} showed the need for personalization in HAR. 
\rem{In their study, models trained only on target user data, outperformed general models that did not train with any target data as well as hybrid models containing some target data during training.}
\add{They trained personal models only on labeled data from the target user as well as hybrid models containing some target data. 
These models outperformed general models that did not see any target data during training.}
\add{If motion patterns between users are different, this follows quite naturally 
from the assumption behind the classification models that the distribution between training and application must be the same.}
\rem{However, the hybrid models still clearly outperformed the unpersonalized models.}
However, because collecting a sufficient amount of data for each target user is not practical, much research focused on methods to fine-tune general models given only a small amount of labeled target data \cite{sani2018matching, parkka2010personalization, hong2015toward, cvetkovic2011semi, reiss2013personalized, sani2017knn, garcia2015building, rokni2018personalized}. 
Acquiring even a small set of labeled data may be costly and impractical. 
Therefore, many researchers also applied unsupervised transfer learning algorithms or semi-supervised learning \cite{ramasamy2018recent, barbosa2018unsupervised, zhao2011cross, lane2011enabling, maekawa2011unsupervised, hachiya2012importance, deng2014cross, wang2018deep, saeedi2018personalized, soleimani2019cross}.
\rem{The problem is also very related to adaptation of sensor placement \cite{chang2020systematic}.
As such, it may be beneficial to look at solutions for one problem to adapt and evaluate elements of it in to the other problem.} 

All this work assumes the availability of the full target user data as one static batch. 
In real applications, however, this data may often not be available.
A new unseen target user would start using the system, and the data would arrive sequentially, presumably until infinity.  
There is little work devoted to personalization in this online setting. 

For example, \cite{cvetkovic2015adapting} present an approach that relies on a general classifier trained on multiple source users and a personal classifier trained on a small subset of the data from the target user. 
During the online phase, a meta-model decides for each incoming instance whether to output the prediction from the general or the personal model. 
Another meta-model decides if the classified instance with its predicted label should be included in the training set of the general model, to be retrained periodically.
Their method assumes the availability of a batch of target user data before the online phase, relies on self-learning and saves all the previous data to retrain the model periodically.   

Similarly, \cite{siirtola2019incremental} use an ensemble trained statically on source individuals as a general model. 
In the online phase, data of the target person becomes available sequentially and is gathered into a small batch containing several activities. 
The general model classifies the batch, and a new ensemble is trained based on these predictions. 
Depending on the confidence of the predictions of the initial base classifier, either the predictions themselves are used as ground truth, or the user is asked for a label. 
Finally, the general and the new ensemble are merged into an overall model. 
Their method relies on a combination of self-learning and active learning. 
Further, personalization does not happen in real-time but is postponed until a batch with data containing several activities has been collected.
In their setting, they do not feature real-time classification either. 
Instead, they divide the incoming stream into chunks, the first one being for personalization only, the next one for classification and testing, the next one again for personalization and so on. 

Moreover, \cite{abdallah2015adaptive} also use an approach that combines active with self-learning and collects a batch of data containing multiple activities. 
Their method clusters the data and extracts cluster features that are passed to an ensemble as input. 
In the online phase, incoming measurements from small batches are clustered and classified.
Personalization takes place through a combination of active and self-learning. 

In \cite{sztyler2017online} a priori data about physical characteristics of the target and source users is used to determine similar source users for a given target.
Then the sensory measurements from the selected source users are taken to train an online classifier.
Their method uses active learning for personalization. 

Quite similarly, \cite{mannini2018classifier} train an incremental SVM on source individuals and update the model incrementally in the online phase, using active learning as well.
They could improve classification accuracy by approximately only 1\%. 

\add{From the related area of activity discovery, \cite{gjoreski2017unsupervised} features an online clustering algorithm which is tailored towards certain human behavioral characteristics.
It can be seen as a personalized approach as it only operates on data from the target user collected during the online phase. 
There is no training phase on source users with different motion patterns. 
Their work is similar to ours in that it works in real time on a new unseen user without requiring any labeled examples of him. 
It differs in that it finds new, unknown patterns over time and groups similar patterns together, instead of assigning them to a closed, pre-defined activity class. 
This feature comes at the inherent drawback of not knowing the activity label for a pattern cluster. Labeling a new, unknown cluster must happen with a user query.} 

All these approaches rely on self or active learning. 
None of \rem{them} \add{the classification algorithms} personalizes in real time or provides a way to balance the stability-plasticity tradeoff.
Our approach, on the other hand, relies on theoretically founded transfer learning that explicitly deals with distribution shifts between training and testing data.
It is fully unsupervised, i.e., no labels of the target user are needed. 
Personalization happens each time right before classifying an activity in real-time. 
Using an exponential average provides a parameter $\alpha$ to adjust the adaptation rate.

\section{Problem Definition} \label{sec:problem_definition}
\begin{figure}
    \centering
    \includegraphics[width=.8\textwidth]{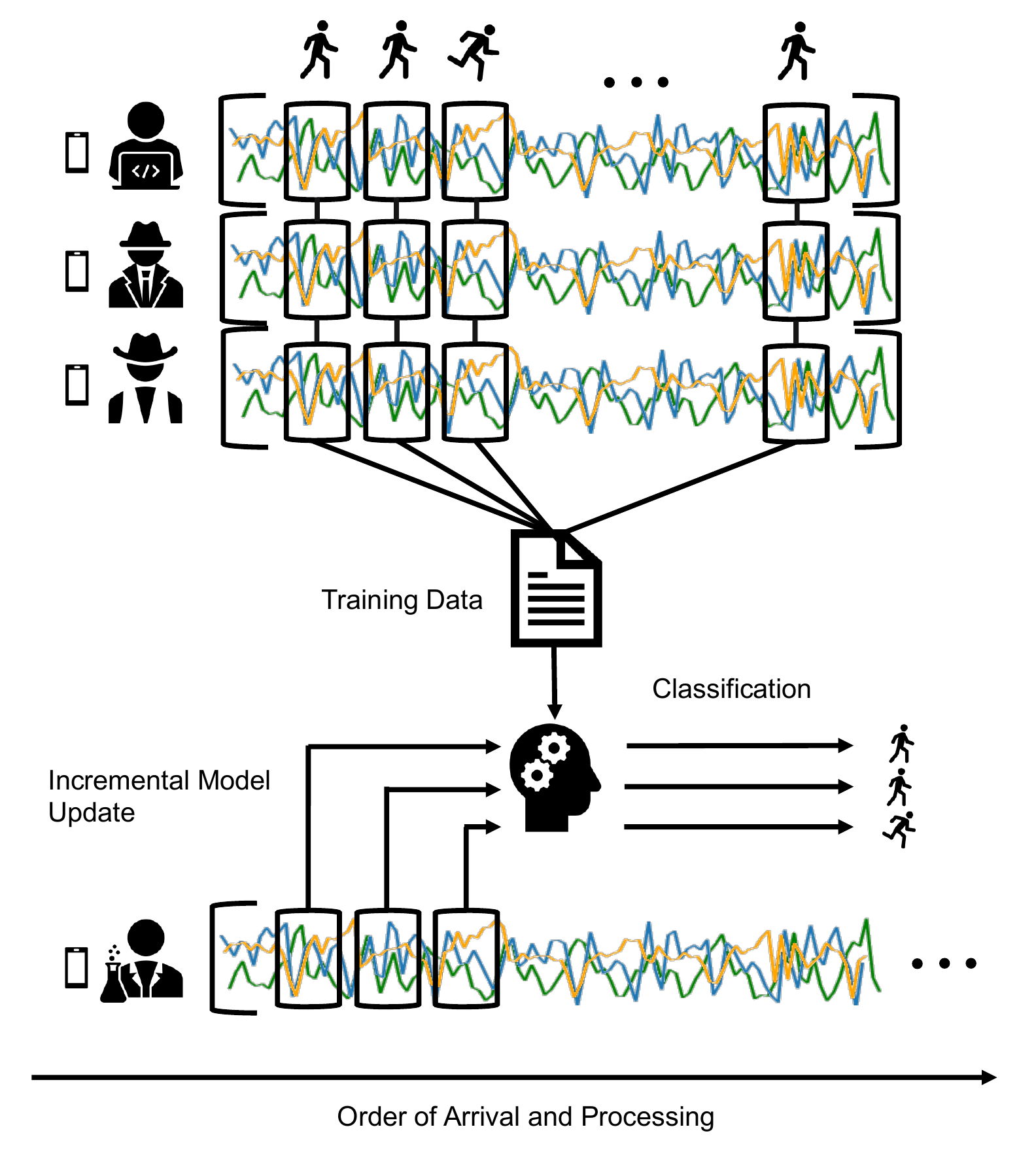}
    \caption{Problem Illustration}
    \label{fig:problem}
\end{figure}

Using a regular smartphone, 3-axis accelerometer measurements \rem{$ x_{i}^{k} = [\hat{x}^{1}_{i}, \hat{x}^{2}_{i}, \hat{x}^{3}_{i}]$} \add{$ x_{i}^{k} = [\hat{x}^{1,k}_{i}, \hat{x}^{2,k}_{i}, \hat{x}^{3,k}_{i}]$} are collected in regular intervals at time step $i$. \rem{$\hat{x}^{1}_{i}, \hat{x}^{2}_{i}, \hat{x}^{3}_{i} \in \mathbb{R}$} \add{$\hat{x}^{1,k}_{i}, \hat{x}^{2,k}_{i}, \hat{x}^{3,k}_{i} \in \mathbb{R}$} denote the respective measured values in the x, y and z accelerometer  dimension for a person $k \in \kappa = \{1, 2 , ..., K\}$.
We assume that person $t \in \kappa$ is the target, and the other ones are source individuals $S = \kappa/\{t\}$. 
For the target person, measurements $x_{i}^{t}$ are subsequently arriving, possibly until infinity $i = \{1, 2, ...\}$. 
For each source person $s \in S$ we assume the availability of a labeled training set \add{ordered over time} of $\hat{N}$ measurements \rem{$\hat{X} =$} \add{$\hat{X}^{s} =$} $\{(x_{1}^{s}, a_{m}),$ $(x_{2}^{s}, a_{m}),$ $...,$ $(x_{\hat{N}}^{s}, a_{m})\}$ with $M$ \add{discrete} activity classes $a_{m} \in A =$ $\{a_{1},$ $a_{2},$ $...,$ $a_{M}\}$. 
Given the time-dependent nature of activities, it is hardly possible to classify an activity based on a single measurement. 
One prominent way of dealing with this problem is to collect the measurements $x_{i}^{k}$ into sliding windows of size $\nu$,  $W_{\tau}^{k} = [x_{\tau c}^{k}, ..., x_{\tau c + \nu -1}^{k}]$, $c$ being the stride between subsequent sliding windows \add{and $\tau$ denoting the $\tau$-th sliding window in order of time.} 
For the target person $t$, $\tau \in \{1, 2, ... \}$  holds. 
The training set \rem{$\hat{X}$} \add{$\hat{X}^{s}$} is transformed into \rem{$X = $} \add{$X^{s} = $} $\{(W_{1}^{s}, a_{m}),$ $(W_{2}^{s}, a_{m}),$ $...,$ $(W_{N}^{s}, a_{m})\}$, \rem{$N = \frac{\hat{N} - w}{c} + 1$} \add{$N = \frac{\hat{N} - \nu}{c} + 1$}, where $a_{m} \in A$ is the most frequent class in the respective sliding window.
So an instance to be classified is not represented by a single measurement but a set of measurements in a sliding window, each measurement representing a feature of the instance. 
The \add{objective} is to classify each subsequent input sliding window $W_{\tau}^{t}$ from the target person by a function $f_{\tau}(W_{\tau}^{k})$ to receive output $a_{m} \in A$ in real time and disregard $W_{\tau}^{t}$  afterwards: $f_{\tau}(W_{\tau}^{k}) = a_{m}$.
\rem{Before} \add{When} classifying $W_{\tau}^{t}$, there is an incremental learning procedure $IL$: $IL(W_{\tau}^{t}, f_{\tau-1}) = f_{\tau}$ that updates $f_{\tau - 1}$ based on that sliding window.  
A supervised machine learning algorithm learns the initial function $f_{0}$ from the \add{whole} training set $X$ \add{$ = \bigcup_{s \in S } X^{s}$      $\forall s \in S $}. Figure \ref{fig:problem} illustrates the setting. 
\add{The following sections explain how the incremental learning procedure $IL$ works and how we learn the initial function $f_{0}$.}

{}

\section{Preliminaries} \label{sec:preliminaries}
\begin{figure}
    \centering
    \includegraphics[width=.8\linewidth]{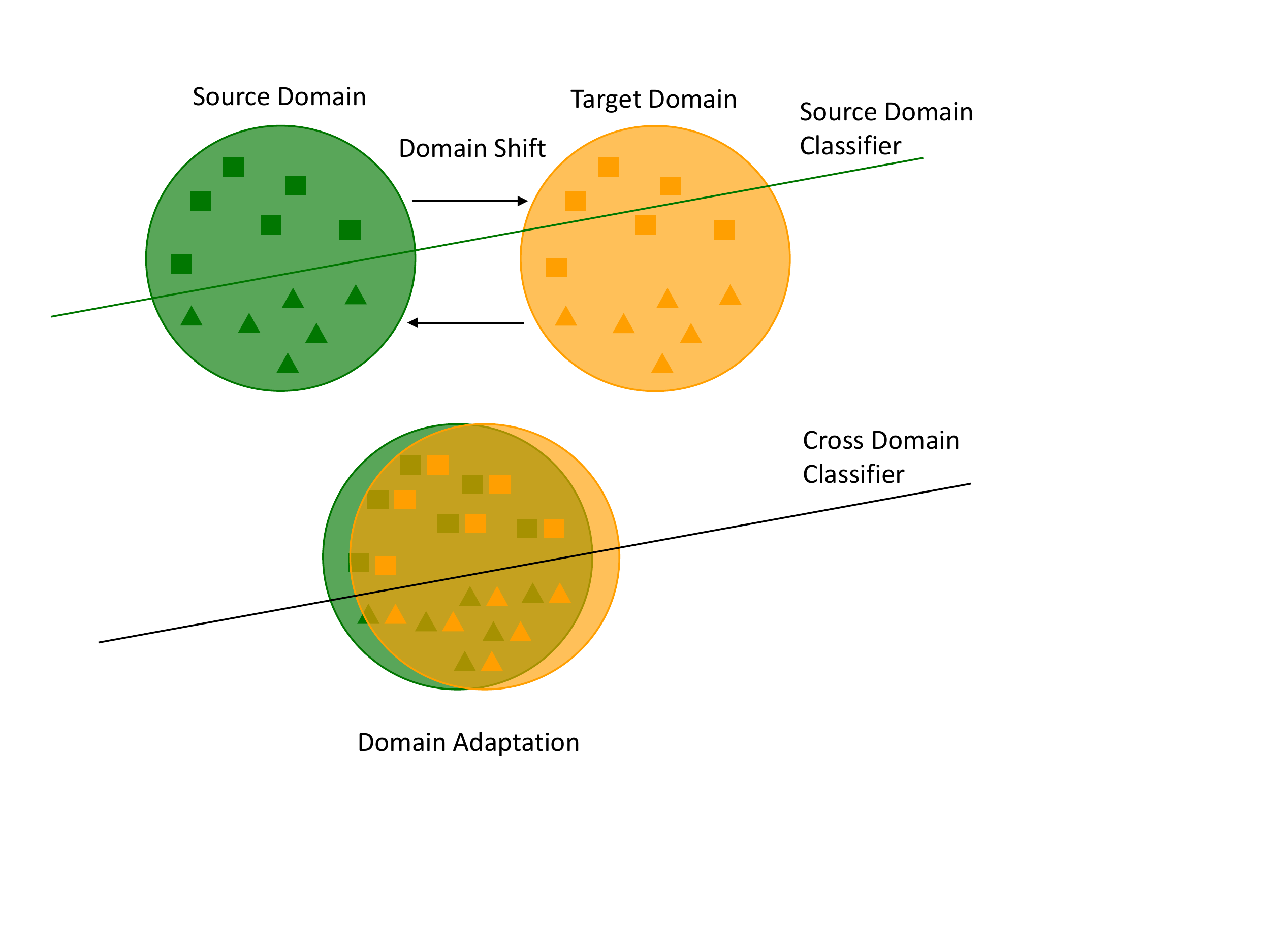}
    \caption{Transformation of source and target into a overlapping feature representation.}
  \label{fig:cross_da}
\end{figure}{}
The approach to be presented is based on \rem{convolutional} neural networks \rem{(CNN)} with an extension of Domain Alignment Batch Normalization (DA-BN) layers  \cite{mancini2018kitting, li2016revisiting}. 
\add{We will first introduce domain adaptation that provides the basic theory for DA-BN.}
We \rem{first} \add{will then} introduce DA-BN layers for the unsupervised batch case, i.e., the test data from the unlabeled target person is available as a single batch at once, and \rem{then} \add{finally} move on to the online case.
Labeled training data from the source individuals is fully available in both cases. 
\subsection{Domain Adaptation} \label{sec:da}
Recent work has demonstrated the effectiveness of DA-BN layers for deep domain adaptation \cite{li2016revisiting, mancini2018kitting, mancini2018boosting, mancini2018robust, cariucci2017autodial, carlucci2017just}. 
Domain adaptation is a branch of transfer learning that deals with problems under the covariate shift assumption.
Given a set of features $\mathcal{X}$ and labels $\mathcal{Y}$ from the same feature/label space for a source domain $\mathcal{S}$ and target domain $\mathcal{T}$, 
the conditional distribution between source and target stays the same, i.e., $P(\mathcal{Y}_{\mathcal{S}} | \mathcal{X}_{\mathcal{S}}) = P(\mathcal{Y}_{\mathcal{T}} | \mathcal{X}_{\mathcal{T}})$, while the marginal distributions differ, i.e., $P(\mathcal{X}_{\mathcal{S}}) \neq P(\mathcal{X}_{\mathcal{T}})$ \cite{pan2009survey}.
DA-BN within a \rem{CNN} \add{neural network} is based on the following results. 
As \cite{shimodaira2000improving} has shown, a learner trained on a given source domain will not work optimally on a target domain when the marginal distributions of the domains are different. 
A solution is to find a feature representation that maximizes the overlap between the domains while also maximizing class separability.
Training a classifier that minimizes the error on the source domain using such a feature representation minimizes the classification error on both source and target domain \cite{ben2007analysis, ben2010theory}. 
As we will see, \rem{CNNs} \add{neural networks} with DA-BN follow this theory. 
Figure \ref{fig:cross_da} is an illustration of this concept. 
The concept is generalizable to the case where \rem{$\mathcal{K}$} multiple 
source domains $\mathcal{S}_{j}$ are present.
\add{We model the personalisation problem in HAR as such a multi source domain adaptation problem. 
Each user $s\in S$ is its own domain $\mathcal{S}_{j}$.
Each set of measurements $W_{\tau}^{s}$ is drawn from $P(\mathcal{X}_{\mathcal{S}_{j}})$ and $(W_{\tau}^{s}, a_{m})$ from $P(\mathcal{Y}_{\mathcal{S}_j} | \mathcal{X}_{\mathcal{S}_j})$. 
The equivalent notions hold for the target user, i.e., $W_{\tau}^{t}$ is drawn from $P(\mathcal{X}_{\mathcal{T}})$ and $(W_{\tau}^{t}, a_{m})$ from $P(\mathcal{Y}_{\mathcal{T}} | \mathcal{X}_{\mathcal{T}})$.
The conditional distribution between all sources $\mathcal{S}_j$ and the target $\mathcal{T}$ is assumed to be equal. 
We further assume that the difference between the motion pattern of users leads to covariate shift.
For a given source user $s\in S$, we assume that his motion pattern does not change over time, i.e., the distribution of sensory measurements of a source user is equal and only differs between source users.
The next sections describe how we align the feature distributions of all users and train a classifier on the aligned sources.}
\rem{In HAR each individual can be seen as its own domain and the differences in the motion patterns between individuals as the covariate shift.}

\subsection{Batch Normalization} \label{sec:bn}
We now review Batch Normalization, before explaining some existing extensions as well as our innovations for domain adaptation.
Batch normalization \cite{ioffe2015batch} is a well-known technique to reduce covariate shift within deep neural network layers and thus to stabilize and accelerate training. 
\add{Let $L$ be the size (number of neurons) of any fully connected hidden layer, and $l \in \{1, 2, ..., L\}$ is the index of a specific channel (neuron) of the layer.}
The idea is to keep the input distribution for a given layer constant by replacing \rem{the channel input $z$ with its standardized value $\hat{z}$, using its mean $\mu$ and variance $\sigma^{2}$} \add{the input $z^{l}$ of channel $l$ with its standardized value $\hat{z}^{l}$ using its mean $\mu^{l}$ and variance ${\sigma^{2}}^{l}$. In the following we omit the channel index $l$ for simplicity.} 
During training in iteration $b$, $\mu$ and $\sigma^{2}$ are computed over the input batch $\mathcal{B}$ of the respective layer. 
\begin{equation} \label{eq:batch_mean}
\mu_{b} = \frac{1}{|\mathcal{B}_{b}|} \sum_{z \in \mathcal{B}_{b}} z 
\end{equation}
\begin{equation} \label{eq:batch_var}
\sigma^{2}_{b} = \frac{1}{|\mathcal{B}_{b}|} \sum_{z \in \mathcal{B}_{b}} (z - \mu_{b})^{2}
\end{equation}{}
An \rem{exponential average} \add{expected value} of these statistics over subsequent batches is computed to be used as a global estimate for $\bar{\mu}$, $\bar{\sigma}^{2}$ in the testing phase.
\add{In line with most implementations and other work on batch normalization, we use an exponentially weighted average to compute the expected value. 
This is done so that the network model is applicable in situations where there is only a single testing instance and thus no batch to compute mean and variance from.}
\begin{equation} \label{eq:exp_mean}
\bar{\mu}_{b} = (1 - \alpha) \bar{\mu}_{b-1} + \alpha \mu_{b}
\end{equation}{}
\begin{equation} \label{eq:exp_var}
\bar{\sigma}^{2}_{b} = (1 - \alpha) \bar{\sigma}^{2}_{b-1} + \alpha \frac{|\mathcal{B}_{b}|}{|\mathcal{B}_{b}| - 1} \sigma^{2}_{b}
\end{equation}{}
\begin{displaymath}
\alpha \in (0, 1) \quad \add{|\mathcal{B}_{b}| > 1}
\end{displaymath}{}

\add{$\frac{|\mathcal{B}_{b}|}{|\mathcal{B}_{b}| - 1}$ in Equation \ref{eq:exp_var} additionally corrects the biased variance estimator from Equation \ref{eq:batch_var}.} 
Irrespective of the normalization happening during training using batch estimates or testing using global estimates, the normalized input $\hat{z}$ is computed by: 
\begin{equation} \label{eq: bn}
    \hat{z} = \gamma \frac{z - \mu}{\sqrt{\sigma^{2} + \epsilon}} + \beta
\end{equation}{}

$\gamma$ and $\beta$ being trainable parameters letting the network shift the imposed distribution.
$\epsilon$ is a (very small) constant for numerical stability. 
\add{Applying Equation \ref{eq: bn} to data distributed with mean $\mu$ and variance $\sigma^2$ makes the data follow a distribution with mean $\beta$ and standard deviation $\gamma$. 
The batch size chosen should be sufficiently large to compute mean and variance estimates.
Since the gradients of the loss w.r.t.\ $\mu$, $\sigma^{2}$, $z$, $\gamma$ and $\beta$ can be computed, $\gamma$ and $\beta$ can be learned, and the error can be backpropagated through the layer during training. 
The respective formalizations are omitted for space considerations and can be found in \cite{ioffe2015batch}}.

\subsection{Domain Adaptive Batch Normalization} \label{sec:da-bn}
\add{Neural network classifiers assume that the marginal and conditional distribution between training and testing stays the same. 
Under covariate shift, this assumption is violated since the marginal distributions are different between training on a source domain and testing on a target domain.
As explained in Section \ref{sec:da}, a sound way of dealing with this issue is to find a feature representation that maximizes domain overlap as well as class separability.
Domain Adaptive Batch Normalization \cite{li2016revisiting} aligns the feature distribution between source and target within the hidden layers. 
This reduces the covariate shift between domains.
It does the alignment by a simple change in the testing phase.}
\rem{Batch normalization can be applied to unsupervised single-source-single-target domain adaptation by a simple change in the testing phase, as proposed by \cite{li2016revisiting}.} 
Instead of applying Equation \ref{eq: bn} with the global estimates $\bar{\mu}$, $\bar{\sigma}^{2}$, target-specific estimates are computed using the fully available unlabeled target test data $\mathcal{X}_{\mathcal{T}}$.
\add{The distribution of the extracted features from the target test data should overlap with the distribution during training.}
\rem{In case of multiple source domains each batch $\mathcal{B}$ must only contain instances from the same source $\mathcal{S}_{j}$ during training.}
\add{The approach can also be used in multi-source domain adaptation.
One must make sure that a given batch $\mathcal{B}$ only contains instances from the same source.}
\add{Therefore, data from different sources will be in different batches.}
\add{Each batch should also be sampled randomly without replacement, so that it contains all classes with the same frequency as in the respective source domain. 
If $|\mathcal{B}| >> M$ and $|\mathcal{B}|$ is sufficiently large, the statistics computed over each batch are representative for the respective source domain. }
So each domain is normalized using its own domain-specific statistics.
This imposes the same target distribution on the features, \add{no matter how far apart the domains are from each other}.
Thus, the feature distribution of the input of the classification layer should overlap for all domains. 
All other parameters (weight and bias terms) of the neural network remain shared.
During training they are optimized to minimize classification loss. 
\add{As there are no changes to the normalization operation during training, backpropagation works as in regular batch normalization.}
\add{The overall result is that,} \rem{As a result} the network learns a feature transformation that maximizes class separability while making the domains overlap.

\subsection{Online Domain Adaptive Batch Normalization} \label{sec:online-da-bn}
The method outlined in the previous section works in the batch setting.
Test data from the target domain $\mathcal{X}_{\mathcal{T}}$ or at least a sufficiently large subset of it is available to estimate the mean and variance specific to the target domain.
\cite{mancini2018kitting} proposed an online extension of DA-BN for visual object recognition under changing visual conditions \add{for single-source-single-target domain adaptation}.
Their method features the following adjustment of the online testing phase of DA-BN. 
From the stream of test data, small batches of the incoming instances (images in their case) are collected.
Equivalently to the training procedure in regular batch normalization, the mean $\mu_{b}$ and variance $\sigma^{2}_{b}$ are computed using Equation \ref{eq:batch_mean} and Equation \ref{eq:batch_var}, and the global estimates $\bar{\mu}_{b}$, $\bar{\sigma}^{2}_{b}$ are updated with Equation \ref{eq:exp_mean} and Equation \ref{eq:exp_var}. 
These global estimates are then used in Equation \ref{eq: bn} to transform $\mathcal{B}_{b}$.
\add{Note that $b$ stands here for iterations during the online phase, not the training phase.}
\rem{Here, $b$ denotes the $b$-th batch to be processed in the online phase.}
After collecting and processing each batch, an incremental adaptation step takes place. 

\section{Incremental Online Domain Adaptive Batch Normalization} \label{sec:approach}
Given our problem formulation, \rem{their solution is} \add{the current solutions are} not directly applicable. 
\add{First, in our scenario there are several sources during training, as opposed to \cite{mancini2018kitting}.
We solve this by using the same strategy to handle several sources as described in Section \ref{sec:da-bn}.
More importantly however, we also need to make a crucial adjustment in the online phase. }
Section \ref{sec:problem_definition} has explained that we collect measurements $x_{i}^{k}$ into sequences of measurements $W_{\tau}^{k}$ (sliding windows). 
However, we represent one instance of our data with one sliding window.
The data units for the subsequent data-processing steps are entire sliding windows, i.e., one window is a data point. 
The measurements are its features. 
Put differently, a sliding window is a point in the multidimensional feature space.
Since in our case the incremental adaptation step takes place after each incoming sliding window and before real-time classification, we must update the global mean and variance estimates $\bar{\mu}$, $\bar{\sigma}^{2}$ in a way that uses a single instance \add{in the online phase}, i.e., $|\mathcal{B}_{b}| = 1$. 
As all previously proposed DA-BN variants (online and offline) assume a batch of instances ($|\mathcal{B}_b| > 1$), we replace Equation \ref{eq:exp_mean} and Equation \ref{eq:exp_var} with an incremental, exponential formulation in our solution. 
This formulation updates the global statistics directly from the input $z$ of a single channel, based on \cite{finch2009incremental}.
\begin{equation} \label{eq:exp_inc_mean}
\bar{\mu}_{b} = (1 - \alpha) \bar{\mu}_{b-1} + \alpha z
\end{equation}{}
\begin{equation} \label{eq:exp_inc_var}
\bar{\sigma}^{2}_{b} = (1 - \alpha) (\bar{\sigma}^{2}_{b-1} + \alpha (z - \bar{\mu}_{b-1})^{2}) 
\end{equation}{}
\begin{displaymath}
\alpha \in (0, 1) \quad z \in \mathcal{B}_{b} \quad |\mathcal{B}_{b}| = 1
\end{displaymath}{}

The \textit{online adaptation momentum} $\alpha$ in Equation \ref{eq:exp_inc_mean} and Equation \ref{eq:exp_inc_var} is a weighting factor. 
By choosing an $\alpha \in (0, 1)$, one can regulate how strong the influence of a new instance should be on the running mean and running variance estimate. 
So it allows to balance the stability-plasticity tradeoff for the given setting. 

\add{Recall that we do not assume the availability of any target information at the beginning of the online phase. 
The initial estimates for the target are average values over the sources. 
It is in the course of the online phase when multiple target instances have been processed that the global estimates converge towards target domain specific values.}

\rem{Note that Equation \ref{eq:exp_var} computes an unbiased variance estimate.
This means, that there is a correction for the bias introduced by estimating a population statistic from a finite sample. 
Since the statistic is simply computed from a batch, a correction term is known. 
On the other hand, Equation \ref{eq:exp_inc_var} does not correct for bias. 
We are unaware of a bias correction term for the \textit{incremental exponentially weighted variance}. 
Yet, we don't expect this to influence our method significantly. 
Sometimes, other works also apply a biased variance computed over a sample in their methods. 
For instance, regular batch normalization does not correct for bias in the training phase, on purpose, to facilitate gradient computation \cite{ioffe2015batch}.}

\add{Putting things together, the final steps of incremental online domain adaptive batch normalization are as follows.} 

\begin{itemize}
    \item \add{\textbf{Domain-Specific Batch Creation}:
    For all sources $s \in S$, the ordered set of user-specific training data $X^{s} \subseteq X$ gets divided into equal-sized, non-overlapping batches $\mathcal{B}_{p}^{s}$ of size $q = |\mathcal{B}_{p}^{s}|$ $\leq |X^{s}|$, by randomly drawing elements from $X^{s}$ into $\mathcal{B}_{p}^{s}$ without replacement.
    $p \in \{1, ..., \frac{|X^{s}|}{q} \}$ denotes the $p$-th batch of source $s$.
    If $q$ is sufficiently large and $q >> M$, each batch will be a statistically representative sample of its respective domain.}
    \item \add{\textbf{Training Phase}: During training in iteration $b$, one batch is passed to the neural network to update the parameters of the network by backpropagation. 
    From the perspective of any channel $l$ in any layer of the network, $z$ is the input of the respective layer when processing the input of a network $W_{\tau}^{s}$.
    Each channel $l$ in each fully connected layer receives the entire processed batch input from the previous layer to compute a batch mean $\mu_{b}$ and batch variance $\sigma^{2}_{b}$ using Equation \ref{eq:batch_mean} and Equation \ref{eq:batch_var}. 
    These estimates are used to normalize the respective dimension $l$ of each layer input $z$ from the respective batch using Equation \ref{eq: bn}. 
    The normalized value $\hat{z}$ is passed to the activation function of the layer to be processed to the next layer. 
    Each channel $l$ in each layer also tracks global running statistics $\bar{\mu}_{b}$ and $\bar{\sigma}^{2}_{b}$.
    They are updated in each iteration $b$ using Equation \ref{eq:exp_mean} and Equation \ref{eq:exp_var}.
    Since each batch is a representative sample of its domain, $\mu_{b}$ and $\sigma^{2}_{b}$ are domain-specific. 
    The normalization reduces the covariate shift between the domains. 
    The global statistics are weighted averages over the batches and are not domain-specific. 
    Their sole purpose is to serve as initial estimates for the testing phase.
    During the backward pass, $\gamma$ and $\beta$ of each channel (Equation \ref{eq: bn}) are updated. 
    At the end of the training phase, the neural network is our final classifier $f
    _0$. }
    \item \add {\textbf{Online Testing Phase}: In the online testing phase, sliding windows $W^{t}_{\tau}, \tau \in \{1, 2, ... \}$ from the target  are processed one after the other in the order of arrival. 
    The incremental learning and classification step per sliding window $IL(W_{\tau}^{t}, f_{\tau-1})(W_{\tau}^{t}) = f_{\tau} (W_{\tau}^{t}) = a$ happens within one pass through the model. 
    The current sliding window $W_{\tau}^{t}$ is passed to the neural network to be processed by the layers of the network. 
    Each channel $l$ in each fully connected layer updates its global statistics $\bar{\mu}_{b}$ and $\bar{\sigma}^{2}_{b}$ using Equation \ref{eq:exp_inc_mean} and Equation \ref{eq:exp_inc_var} from the layers' input $z$.
    This is the incremental learning procedure $IL$. 
    The updated global statistics in Equation \ref{eq: bn} are used to normalize the respective input $z$.
    $\hat{z}$ gets passed into the layers activation function and forwarded into the next layer.
    The neural network model outputs a class from the input of its final classification layer. }
\end{itemize}

\rem{For the general model $f_{0}$, we train a convolutional neural network (CNN) with online DA-BN layers. 
As described in Section \ref{sec:preliminaries}, these layers perform an incremental learning step $IL$ to subsequently adjust the global model to the target person in the online testing phase. 
As usual, the CNN consists of 3 parts: (1) a regular convolutional block with several convolutional and maxpooling layers, (2) a fully connected block with one or multiple fully connected, online DA-BN layers and (3) a softmax classification layer.}

\section{Experiments} \label{sec:experiments}
\subsection{Experimental Setup}
 
\subsubsection{Dataset}
In our experiments, we use the WISDM dataset publicly available in the UCI Repository \cite{weiss2019smartphone}.
The dataset contains accelerometer and gyroscope measurements collected at approx.\ 20 Hz with a smartphone and a smartwatch from 51 subjects.
It contains data on 18 activities of daily living.
During collection, all subjects had the smartphone in the same pocket and in the same orientation. 
For each subject and activity approx. 3 minutes have been recorded. 
The activities have been recorded separately. 
This means that each person performs one activity for approx. 3 minutes in a row, followed by the next activity etc. 
When looking at the timestamps, one can see that the transitions from one activity to the next one are not continuous, but recording has happened in isolation. 
Also, the data from the smartphone and smartwatch is not synchronized i.e. they have not been collected in parallel. 

\subsubsection{Preprocessing}
In line with most other work on personalized HAR we only use the data from the smartphone accelerometer.
The accelerometer is the most meaningful sensor for motion based HAR.
\rem{We also don't want to influence our results with the effects of sensor fusion.} 
We consider the activities \textit{walking, jogging, walking stairs, sitting} and \textit{standing}.
\add{The other activities, such as \textit{brushing teeth}, are hand-oriented and thus not detectable from a smartphone alone.
Since the smartphone and smartwatch data has not been collected synchronously but in isolation sensor fusion is not possible.}
\add{The selected activities are also the most common ones in related work.} 
Because subjects 09, 16 and 42 did not contain all relevant activities, we disregard their data.
Additionally, as \cite{burns2020personalized} reported issues with the collected measurements of 37 to 40 we disregard them as well. 
As such, we are left with 44 subjects. 

The recording of measurements happened in isolation. Therefore, we group the data by activity and person id for the following preprocessing steps. 
We resample the data so that the sampling frequency is at exactly 20 Hz. 
We also truncate the last measurements to ensure an equal number of measurements per activity and person. 
This yields a perfectly balanced dataset.
We apply a non-centered moving average filter of size 4 for consistency with the online setting. 
The value of a filtered measurement should not be based on future values but should only use measurements from the past. 
Therefore, the average value at timestep $i$ was computed as the average of the last $i$ to $i-3$ values. 
This filter size was chosen as a combination of results from preliminary experiments and common practice in related work \cite{saeedi2018personalized, bruno2013analysis}. 
We apply a min-max normalization with a min-max range of [-78, 78] based on the value range of the accelerometer. 
Finally, sliding windows are of size 40 (2 sec) with 50 \% overlap. 
The size and overlap have been chosen based on an empirical study that has tested HAR models with varying sliding window sizes and overlaps \cite{banos2014window}.
One advantage of using a neural network based model is that feature extraction is part of the overall learning process.
As such we do not extract any hand-crafted features from the sliding windows. 

All in all, we end up with 3560 measurements (2:58 min) separated into 177 3-dimensional sliding windows of size 40 per activity and person for 5 activities and 44 individuals. 
Thus we have $177 * 5 * 44 = 38940$ instances in total. 

\subsubsection{Evaluation Method}
To show the effectiveness of DA-BN layers for personalization in HAR, we conducted several experiments.
\add{They either are baseline experiments to make comparisons to the case when there is no domain adaptation, or they evaluate the domain adaptation method under different circumstances such as the classic batch case or the online case.}
\rem{They evaluate the method in the batch and the online setting.}

Unless otherwise stated, we employ the \textit{leave-one-person-out-cross-validation} (LOPOCV) evaluation model. 
We create $K$ folds and assign the data of each person $k$ to exactly one fold \cite{jordao2018human}. 
So each person is once the target person $t$, while the base model is trained on all the remaining individuals. 
For each fold the classification accuracy is computed. 
In the evaluation section, results are often summarized as medians or means over all folds.  
To make sure that results are comparable, each experiment is based on the same \rem{CNN} \add{network} architecture, with the same hyper parameters and initialization weights.

We conduct the following experiments:
\begin{itemize}
    \item \textbf{Baselines}: First, we create a \textit{Lower Baseline} and an \textit{Upper Baseline}.
    The \textit{Lower Baseline} consists of a regular CNN with regular batch normalization layers in the fully connected block evaluated under the LOPOCV. 
    The \textit{Upper Baseline} uses a different evaluation model than the other experiments. 
    The data of each person is randomly split into a training and a test set. 
    For each person $k \in \kappa$ a regular \rem{CNN} \add{model} is trained and evaluated on $k$s data only. 
    \rem{Results are often summarized as medians or means over all individuals.} 
    \item \textbf{Unsupervised Batch}: A \rem{CNN} \add{network} with DA-BN layers \add{as described in Section \ref{sec:da-bn}} in the fully connected layers is trained. 
    \add{For the test phase,} the held out data of the target person is randomly split into a pre-estimation and test set with varying relative sizes from 1:90 to 9:1. 
    \add{In contrast to the description in Section \ref{sec:da-bn}}, the pre-estimation set, \add{and not the test set,} is passed to the model to estimate the global means $\bar{\mu}$ and variances $\bar{\sigma}^{2}$ \add{for the target person, using Equation \ref{eq:batch_mean} and Equation \ref{eq:batch_var}}. \rem{over the entire batch}
    These estimates are used in Equation \ref{eq: bn} when classifying the test set. \add{This is a more difficult scenario since the dataset to estimate the BN statistics is seperate from the dataset for evaluation.}
    \item \textbf{Supervised Batch}: This experiment is like the \textit{Unsupervised Batch} experiment, except that the pre-estimation set contains labels that are used to additionally tune the network weights for 10 epochs.  
    To allow comparisons, we also tune the weights of the \textit{Lower Baseline}, dubbed \textit{Supervised Baseline}. 
    \item \textbf{Online Unrandomized}: A \rem{CNN} \add{network} with \add{incremental} online DA-BN layers in the fully connected block is trained. 
     \add{This is the method proposed in this work (Section \ref{sec:approach}).}
    The order of sliding windows in the held out data of the target person is kept as in the original dataset. 
    So, all instances of one class are processed before all instances of the next class. 
    The online adaptation momentum $\alpha$ is varied between [0.0001, 0.005]. 
    Instances are processed one at a time, i.e., not as a batch. 
    \item \textbf{Online Randomized}: This experiment is like the \textit{Online Unrandomized} experiment, but the order of the instances is randomized. 
    So activities are uniformly distributed in time. 
    This should simulate a slightly more realistic scenario than keeping the order in blocks of activities, as provided by the authors of the dataset. 
    This experiment is repeated 5 times, varying the order of sliding windows, and results are averaged. 
    We vary the online adaptation momentum $\alpha$ between [0.001, 0.05]
\end{itemize}{}
\add{To sum up, the \textit{Online} experiment uses our approach as proposed in Section \ref{sec:approach}, a \textit{Batch} experiment uses a static variation of the approach described in Section \ref{sec:da-bn}, a \textit{Supervised} experiment additionally uses weight tuning with target labels, and \textit{Baselines} do not use any domain adaptation.}

\subsubsection{Implementation}
\add{For our experiments we trained a convolutional neural network (CNN) architecture with a regular convolutional block and a following fully connected block.}
The CNN model employed in all our experiments consists of 5 1D-convolutional layers with 64 feature maps, a convolutional kernel of size 5, stride 1 and ReLU activation function. 
Zero padding is applied to keep the size of the feature maps constant throughout the convolutional block. 
After the last convolutional layer 1D, non-overlapping max pooling with a kernel size of 4 is applied. 
The following fully connected block consists of 1 fully connected layer with 256 neurons, the batch normalization variation of the respective experiment, a 50 \% dropout rate and a ReLU activation function. 
The classification layer uses a Softmax activation function. 
\rem{Figure \ref{fig:cnn} summarizes the architecture with the respective hyper-parameters.}
As the loss function we have chosen the cross-entropy loss.
During training we employ the ADAM optimizer with a 0.0001 learning rate and 0.001 decay, training for 649 epochs on batches of size 177. 

We determined these hyper parameters to work best in a grid search for a general recognition model. 
We also conducted a grid search for the personal models of the \textit{Upper Baseline}. 
However, the results with the best \textit{Lower Baseline} hyper parameters in the \textit{Upper Baseline} experiment were only slightly different. 
So, for higher comparability, we also employ the same hyper parameters for the \textit{Upper Baseline}, except for the number of epochs. 
We determined these separately for each personal model using early stopping on a validation set. 

The code for the experiments is in Python, using the PyTorch (with CUDA), Numpy and Pandas libraries. 
The experiments have been run on the Pittsburgh Super Computer with NVIDIA Tesla V100 16 GB memory GPUs \cite{Nystrom:2015:BUF:2792745.2792775}.

\subsection{Results}
\subsubsection{Comparison across All Experiments} \label{sec:fst_res}

\begin{figure}
    \centering
    \includegraphics[width=\linewidth]{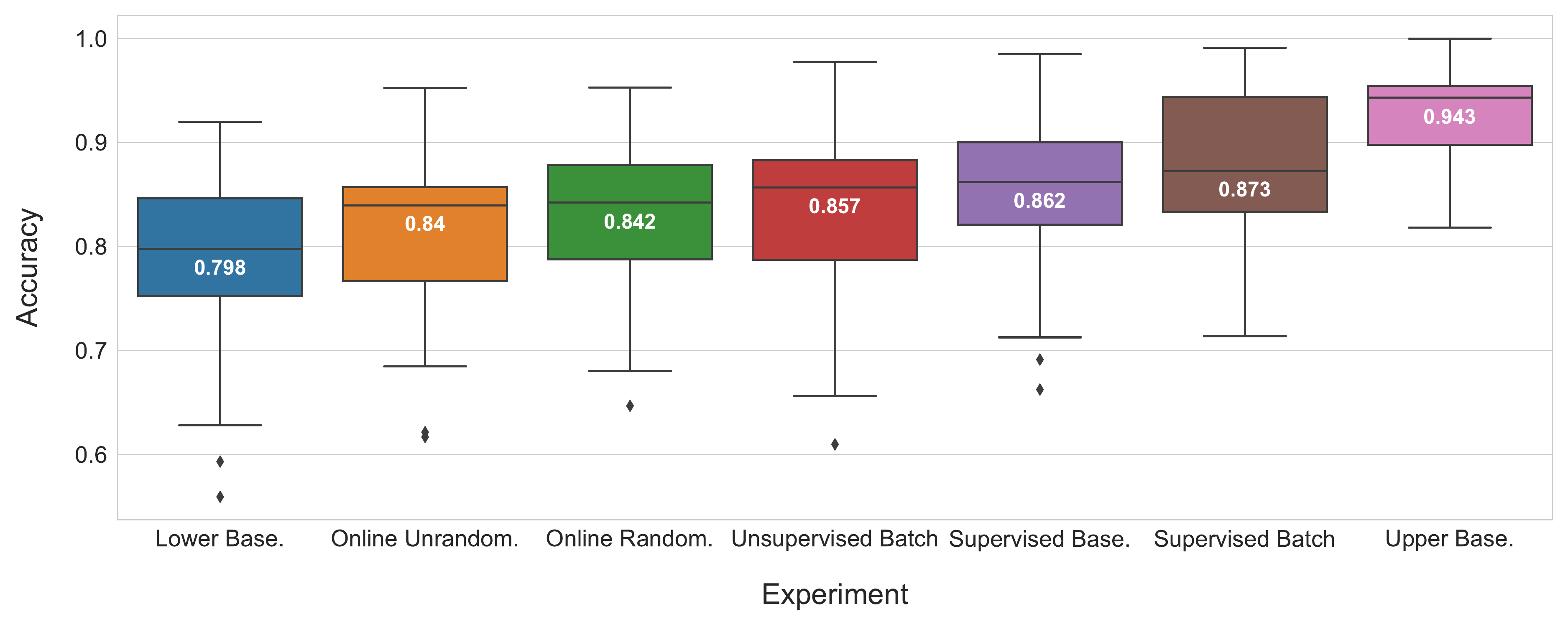}
    \caption{Boxplot summarizing accuracies on all experiments.} 
    \label{fig:boxplot}
\end{figure}

Figure \ref{fig:boxplot} compares the LOPOCV results of the experiments. 
The values in the boxes are the median accuracies. 
For \textit{Online Unrandomized} and \textit{Online Randomized}, the results are for online adaptation momentum values of 0.0009 and 0.01 respectively. 
For \textit{Unsupervised Batch}, \textit{Supervised Baseline} and \textit{Supervised Batch}, 10 \% of the target data is used as the pre-estimation set and the remaining 90 \% are used for testing. 
    
As intended and expected, the \textit{Upper Baseline} sets the maximal possible detection accuracies through personalization. 
On the other hand, the \textit{Lower Baseline} must be improved upon for personaliztion to be any good. 
The (unsupervised) online experiments have slightly smaller improvements over the \textit{Lower Baseline} than the \textit{Unsupervised Batch} experiment. 
The supervised (batch) experiments outperform all the unsupervised ones. 
The boxplot contains outliers towards the lower tail, for most but two experiments. 
These are users for whom the accuracy (``user performance'' in the following) is very low compared to the other users. 

Comparing the unsupervised experiments to the \textit{Lower Baseline}, we see an improvement over the whole distribution of users: 
The medians, the Inter Quartile Range (IQR) and the minimum and maximum are higher, as well as the two outliers (User 10 and 14).  
The biggest improvement can be seen between the minimum of the \textit{Lower Baseline} and the \textit{Online Unrandomized} experiment. 
This suggests that in particular users with low accuracies on the \textit{Lower Baseline} experience significant improvements. 
The same effect can be seen when comparing the accuracies of the outliers on the \textit{Supervised Baseline} to the \textit{Supervised Batch} case.
For some users, accuracies already are above 90 \% using a simple general model.
So it is important to improve detection accuracy for users who are different and hard to classify by the general model. 
Our approach seems to do this, as we will see later when discussing Figure \ref{fig:barchart} and Figure \ref{fig:person}.

Between the \textit{Online Unrandomized} and the \textit{Online Randomized} experiment, median, maximum and minimum are almost equal. 
User 10 and 14 are doing worse in the \textit{Online Unrandomized} case, pushing its overall results a little bit down. 
Although, the IQR is of the same size, its upper border is slightly higher for the \textit{Online Randomized} experiment. 
This suggests that accuracy for more users is higher in the randomized case.
A reason could be that in the unrandomized case all instances of one activity are processed before all instances of the next one. 
The DA-BN layer needs to process some instances of the new activity to adjust its statistics to the new pattern; this might lead to an artificial ``concept drift''. 
During that time, instances of the new activity are classified using statistics based on the previous class. 
In the randomized case, this does not happen. 
As this data is randomized, concept drift occurs only once at the beginning when data of the new user arrives. 
The statistics converge towards their target values and don't change much until the end of the online phase. 
Nevertheless, both cases show strong improvements over the \textit{Lower Baseline} with result not too different from each other. 
It shows how online DA-BN is applicable in different scenarios. 
We will see in Figure \ref{fig:momentum} that this has to do with the choice of online adaptation momentum $\alpha$. 

When comparing the \textit{Online Randomized} to the \textit{Unsupervised Batch} experiment, one can see that the minimum accuracy and one of the outlier's accuracy are lower for the  \textit{Unsupervised Batch} case. 
It means that the lowest performers are doing better in the online case than in the batch case. 
However, this might be a random effect only happening on the two lowest performers on this specific data.
The median, the 75th percentile and the maximum are higher, suggesting that accuracy for the average and top performers is higher in the batch case.

The \textit{Supervised Baseline} improves the median accuracy over the \textit{Lower Baseline} by 6\%. 
That is the effect of tuning the network weights with a small amount of labeled target user data. 
However, the difference between the median and the maximum of the \textit{Supervised Baseline} to the \textit{Unsupervised Batch} experiment is only marginal. 
This shows the strength of our approach in the batch setting.
It means that the results of our unsupervised approach are not much worse than the results with supervised fine-tuning. 
Nevertheless, supervised fine-tuning obviously beats an unsupervised approach. 
We can see that looking at the higher minimum value and the thinner, upward shifted IQR of the \textit{Supervised Baseline}. 
Still, applying DA-BN on top of weight tuning improves activity recognition. 

\subsubsection{Average Results for Groups of Users} 

\begin{figure}
    \centering
    \includegraphics[width=\linewidth]{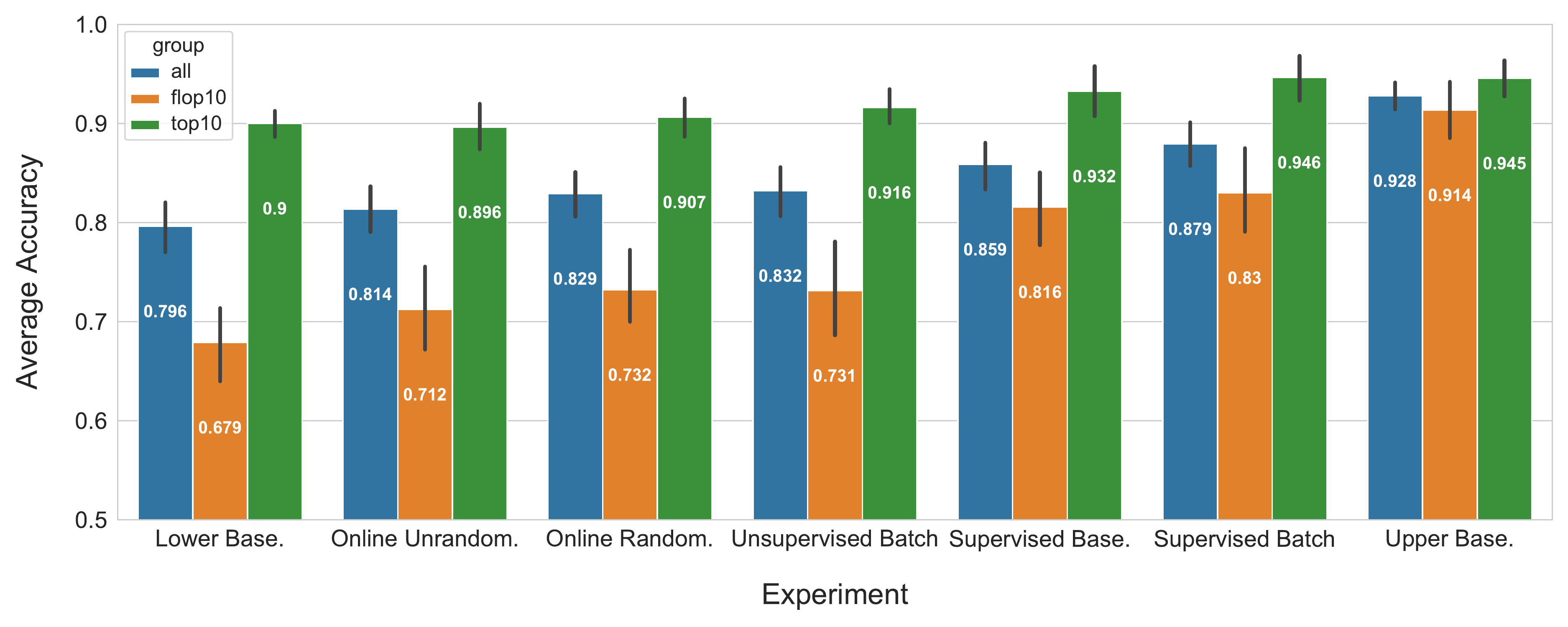}
    \caption{Average accuracies for all experiments.}
    \label{fig:barchart}
\end{figure}

Figure \ref{fig:barchart} compares the average accuracies over all users, as well as results on a subset of the 10 best (dubbed ``top 10'') and worst (``flop 10'') users on the \textit{Lower Baseline}. 
As in Figure \ref{fig:boxplot}, we report results for \textit{Online Unrandomized} and \textit{Online Randomized} with a online adaptation momentum of 0.0009 and 0.01 respectively. 
Equally, for \textit{Unsupervised Batch}, \textit{Supervised Baseline} and \textit{Supervised Batch} 10 \% of the target data is used as the pre-estimation set and the remaining 90 \% is used for testing. 

When looking at the results for all users, there is almost no difference between the \textit{Unsupervised Batch} and the \textit{Online Randomized} experiment, while there is a 1 \% difference when comparing \textit{Online Randomized} and \textit{Online Unrandomized}. 
This again, comes from the relatively good performance of User 10 and User 14 in the \textit{Online Randomized} case. 
Their relatively worse performance in the other two cases pushes the respective averages down.
Therefore, this same pattern can be seen for the flop 10 but not for the top 10. 

For the \textit{Supervised Batch} case, the improvement on all users over the \textit{Supervised Baseline} is higher in terms of mean than median. 
This is because the \textit{Supervised Batch} sharply improved the outliers. 
It suggests that tuning weights in connection with DA-BN layers is even more beneficial in the supervised than in the unsupervised case. 

We also see that using DA-BN has the greatest effect on the flop 10.
There is a big leap from the \textit{Lower Baseline} to the \textit{Online Unrandomized} case.
However, the improvement from adding supervision is obviously larger than from unsupervised DA-BN. 
DA-BN also improves in the supervised case but not as much as in the unsupervised (online) cases.
For the top 10, the unsupervised online experiments do not show an improvement. 
\textit{Unsupervised Batch} and \textit{Supervised Batch}, however, still show an improvement of approx. 2 \% over their respective baseline.  
\textit{Supervised Batch} even achieves results equal to the \textit{Upper Baseline}.

All in all, using DA-BN consistently improves detection accuracy, be it in the online or batch, supervised or unsupervised case. 
For users who do not perform well under a general model there is more room for improvement and a higher overall effect. 
For the top 10 there is not so much room for improvement. 
Still DA-BN shows a significant effect. 
There is a difference in performance in the online randomized and unrandomized case that warrants further investigation. 

\subsubsection{Online Randomized Improvement per Person}

\begin{figure}
    \centering
    \includegraphics[width=\linewidth]{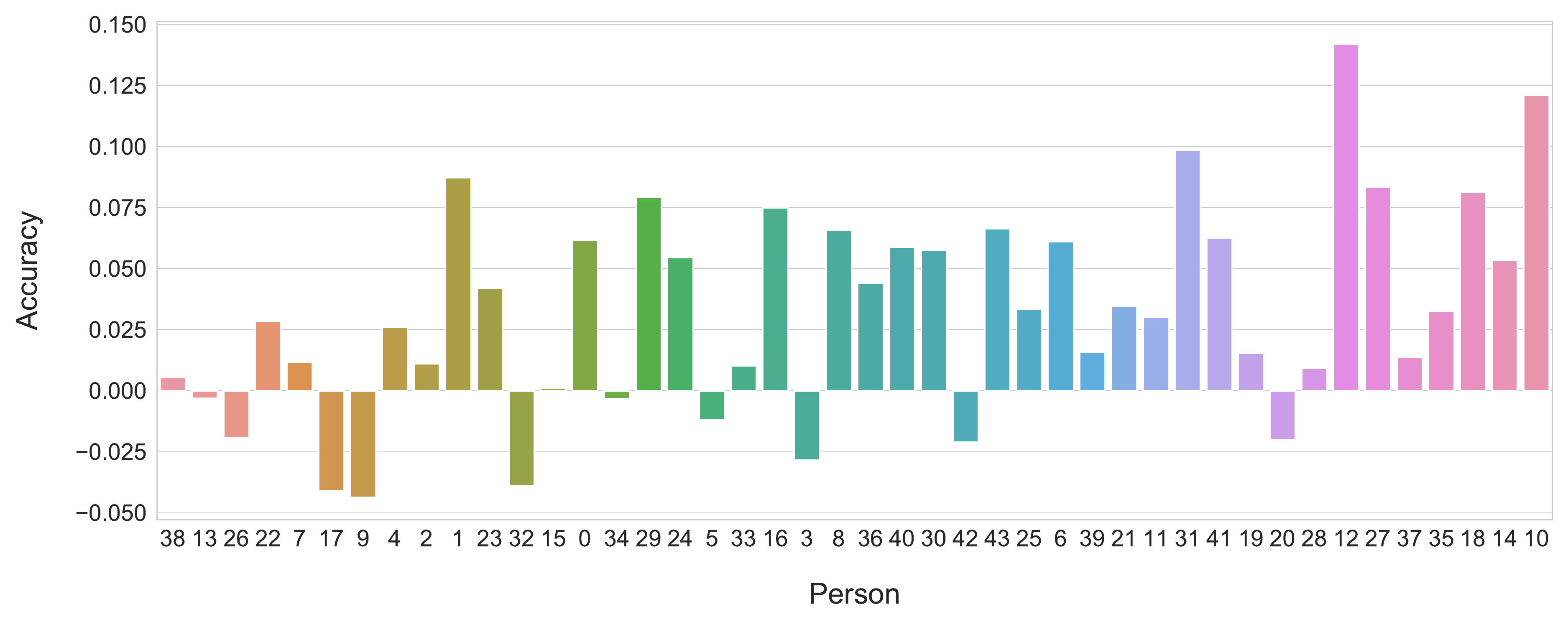}
    \caption{Improvement for each person in the \textit{Online Randomized} experiment with online adaptation momentum of 0.01 over the \textit{Lower Baseline}.}
    \label{fig:person}
\end{figure}

Figure \ref{fig:person} shows the accuracy improvement for the \textit{Online Randomized} experiment over the \textit{Lower Baseline} for each user.
The x-axis is in descending rank order based on the users \textit{Lower Baseline} accuracy. 
To illustrate, User 38 has the highest accuracy on the \textit{Lower Baseline}. 

There are few individuals for whom the accuracy goes down.
The biggest decline is about 4 \%. 
It seems that users who are performing well on the \textit{Lower Baseline} are more likely to experience a decrease in performance. 
Other studies also showed decreased performance for some users after personalization, cf.\ \cite{qin2019cross, chang2020systematic, deng2014cross}. 
In transfer learning this effect is known as negative transfer. 
Recall from Section \ref{sec:preliminaries} that domain adaptation assumes the conditional distribution between source and target to stay the same, while the marginal distributions are different. 
In this real world scenario, the assumption of equal conditional distributions, i.e. $P(\mathcal{Y}_{\mathcal{S}} | \mathcal{X}_{\mathcal{S}}) = P(\mathcal{Y}_{\mathcal{T}} | \mathcal{X}_{\mathcal{T}})$, may be violated between some individuals, possibly leading to the negative transfer.
However, for most users the accuracy improves with gains of up to 14 \%. 
Some of the top gainers, namely Users 10, 12, 18, and 27, are among the lowest performers on the \textit{Lower Baseline}. 
In fact, User 10 has the lowest accuracy on the \textit{Lower Baseline} and is the second biggest gainer.
This is in line with what we have seen in Figure \ref{fig:barchart}, however we had expected this relationship to be stronger. 
For instance, User 1 has the 10th best accuracy on the \textit{Lower Baseline} but has the 4th highest improvement. 
This makes him the top performer of the \textit{Online Randomized} experiment. 

\subsubsection{Impact of Pre-Estimation Set Size}
\begin{figure}
    \centering
    \includegraphics[width=\linewidth]{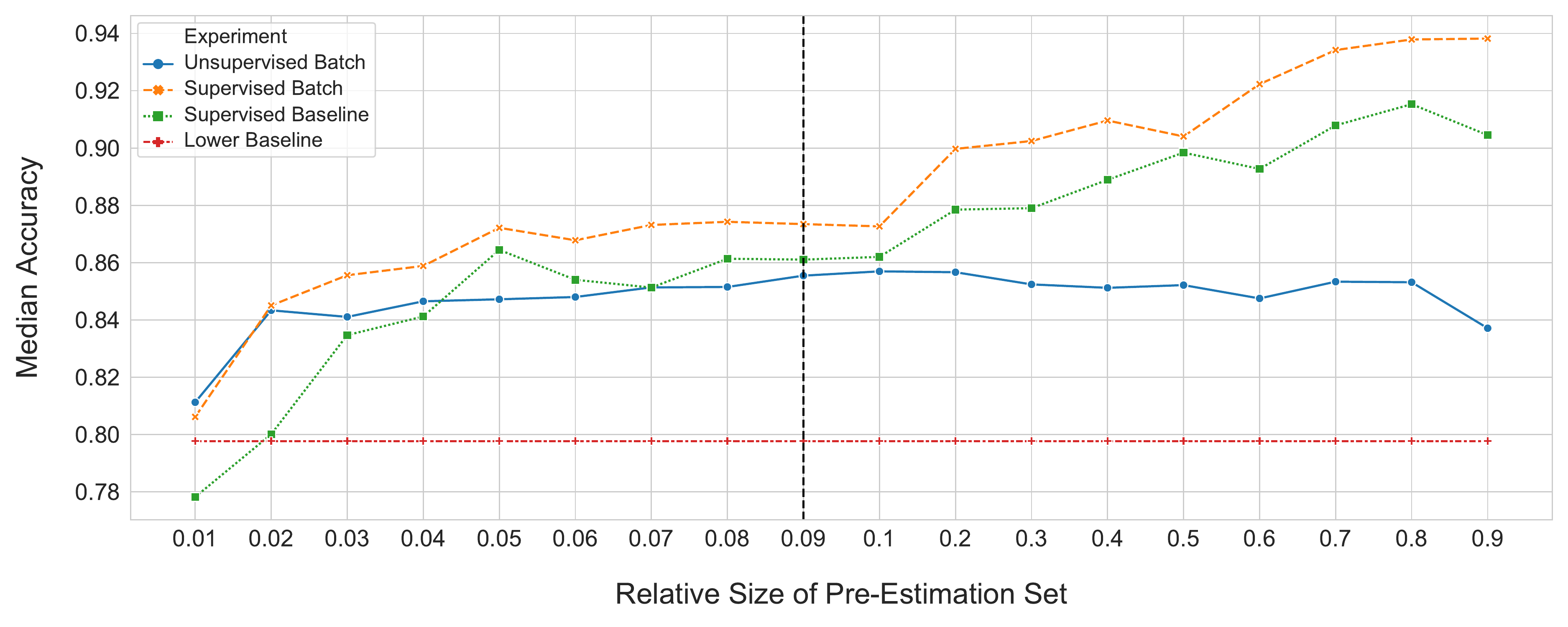}
    \caption{Median LOPOCV results of the batch experiments depending on the target persons relative pre-estimation set size.}
    \label{fig:lineplot}
\end{figure}

Figure \ref{fig:lineplot} displays the median LOPOCV accuracy depending on the relative pre-estimation set size. 
The x-axis denotes the relative size of the pre-estimation batch compared to the overall size of the target data. 
Note that the scale of the x-axis is not linear.

All experiments start with a sharp increase in accuracy. 
The \textit{Supervised Batch} already improves accuracy by 1 \% over the \textit{Lower Baseline} to 0.81, given only 1 \% (8 instances / 9 sec) of the data of the target individuals for pre-estimation. 
Given 2 \% of this data, accuracy sharply increases to 0.84. 
From there on, accuracy slowly goes to almost 0.86 with 10 \% of the target person's data. 
These results show how DA-BN can achieve strong improvements with only little data from the target person and without any label. 
They also indicate how adaptive the online algorithm should be in the given setting. 
When the online algorithm has processed 16 sliding windows (x = 0.02 / 17 sec) the DA-BN statistics could actually be predominantly based on data from the target person only. 
This speaks for a rather high online adaption momentum in the beginning.
Over time the gain due to fast adaption decreases. 
In this case it might be beneficial to have a lower online adaptation momentum to be more robust towards, say, unbalanced class distributions, short within user temporal changes, noise, etc. 
Developing a method that continuously adjust the online adaptation momentum on its own during the online phase might be a promising future direction. 

On the other side, using only a small subset of a target person's labeled data to fine tune the network weights (without DA-BN) has a negative impact on accuracy. 
It drops by 2 \% under the \textit{Lower Baseline}. 
Until x = 0.05 the \textit{Unsupervised Batch} performs better than the \textit{Supervised Baseline} and is only marginally better until x = 0.2. 
From there on, the advantages of supervised fine-tuning come to bear. 
The \textit{Supervised Baseline} becomes better than the \textit{Unsupervised Batch}. 
Comparing these results to the \textit{Supervised Batch}, one can see that it is consistently better than the \textit{Supervised Baseline} by 1 to 4 \%. 
Using DA-BN in the supervised setting mitigates the initial loss at x = 0.01 and already improves over the \textit{Unsupervised Batch} very early at x = 0.02. 
Thus, it is always beneficial to apply DA-BN. 

\subsubsection{Impact of Online Adaptation Momentum} 

\begin{figure}
    \centering
    \includegraphics[width=\linewidth]{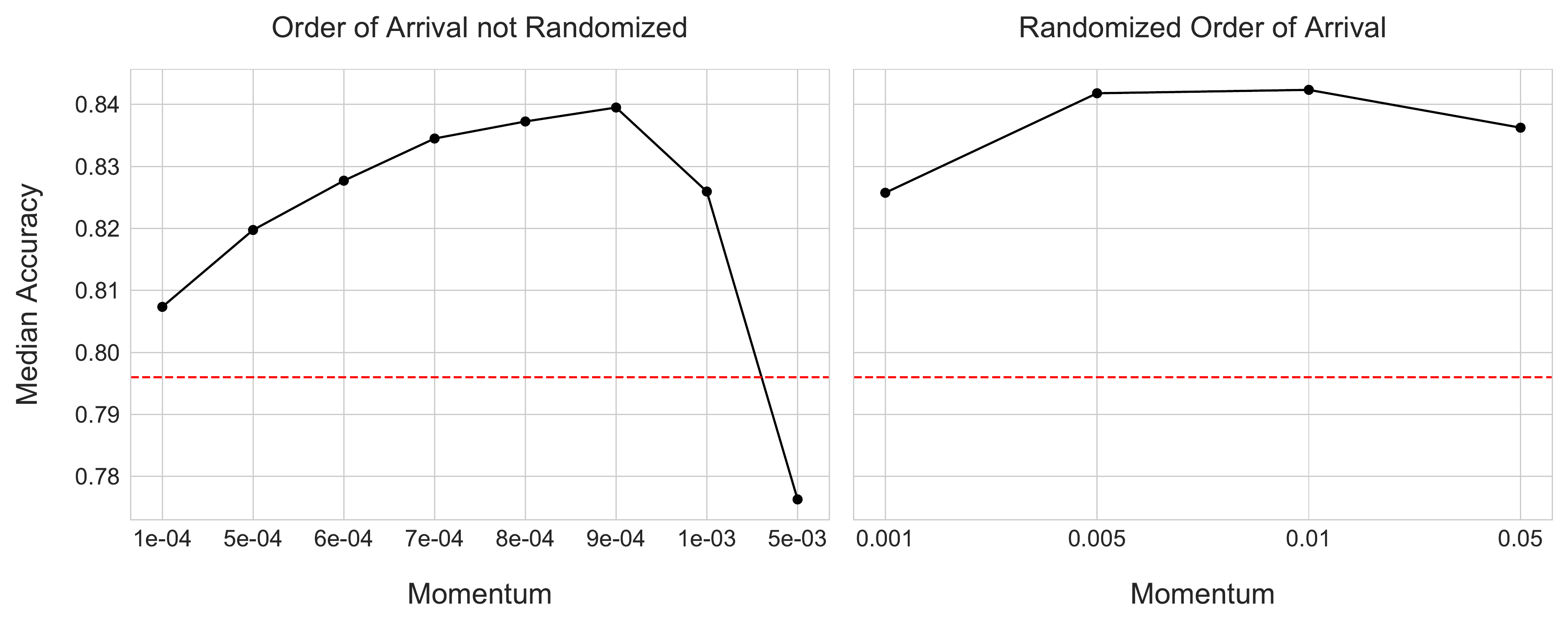}
    \caption{Median LOPOCV results of the online experiments depending on the online adaptation momentum. 
    The red dotted line corresponds to the median \textit{Lower Baseline} performance.
    Note that the scale of the x-axis is not linear.}
    \label{fig:momentum}
\end{figure}

Operating in a dynamic online environment leads to the problem of stability-plasticity that incurs a tradeoff between the ability to take in new knowledge and "forget" old information \cite{gepperth2016incremental}.
Finding an appropriate way to regulate this tradeoff is key in optimizing online detection performance.
In contrast to other online personalization approaches in HAR, online DA-BN can explicitly regulate this balance using the online adaptation momentum. 
Figure \ref{fig:momentum} shows that, depending on whether we are looking at the \textit{Online Randomized} or \textit{Online Unrandomized} case, very different values for that hyper parameter are optimal. 
In the \textit{Online Unrandomized} case, activities arrive in blocks, one after the other. 
Thus, using rather high momentum values confines the DA-BN statistics to the pattern of one activity only. 
This has a negative impact on detection accuracy. 
On the other hand, in the \textit{Online Randomized} case, activities do not arrive in blocks but are mixed. 
Estimating the DA-BN statistics from the last, say, 16 sliding windows reflects the overall pattern across all activities better. 
In this case, being more adaptive and ``forgetting'' the statistics over all users in favor of the user-specific statistics leads to better detection accuracies. 
We have discussed this effect when we were presenting Figure \ref{fig:boxplot}.
We can see how crucial it is to regulate the strength of adaptation depending on the setting in online HAR. 

\section{Discussion} \label{sec:limitations}
\add{ 
In the following we discuss this work and some future work.

We have outlined that in the online machine learning setting it is important to be able to regulate the stability plasticity trade-off. 
The current method allows regulating the trade-off using online adaptation momentum $\alpha$.
Our last experiment confirms that importance. 
We have seen that the optimal momentum is different for the \textit{Online Randomized} and \textit{Online Unrandomized} case.  
However, it comes with the drawback of choosing an appropriate value for the given data. 
Unfortunately, currently there is no way to choose an appropriate value besides a hyper parameter search. 
Further, the momentum in this work has been a static, fixed parameter across all subjects and across time. 
Developing an algorithm that dynamically adapts the momentum in the online phase might be an essential improvement.
}

\add{
The experiments have further shown that there are individuals for whom the adaptation method works better than for others. 
For some subjects the accuracy even decreased. 
This has most probably to do with the covariate shift assumption (outlined in Section \ref{sec:da}). 
The validity of this assumption in HAR dictates how well DA-BN can  be used for personalization or other related adaptation problems (e.g. sensor placement adaptation).
Human motion patterns are very complex, and the statistical assumptions underlying each problem are neither clear nor consistent between individuals, activities, sensors or sensor placements. 
One big area for future work is to empirically study the method on different datasets.  
This could include datasets that contain very diverse subjects (e.g., elderly persons as well as children), different sensors (e.g., watch) and sensor placements (e.g., left pocket, right pocket, different orientations within a pocket), more and more complex activities and datasets collected in real-life scenarios over long periods of time.} 

\add{
When thinking of applying the approach to other activities, the question of scalability with the number of activities arises.
In audio- or vision-based activity recognition, we see neural models with over 500 activity classes. 
In other areas such as image classification, it is not uncommon to have 20,000 classes. 
One might be tempted to think that DA-BN simply scales as the neural architecture does. 
Recall from Section \ref{sec:preliminaries} that during training the batches should be representative for their respective domain, i.e., they should contain each class with the same frequency as the entire domain. 
In the predominant case where the batch size is much bigger than the no. of activities ($|\mathcal{B}| >> M$), randomly selecting a sufficiently large number of samples from a domain into the batches leads to the desired effect.
However, when the number of classes becomes very large, one cannot simply increase the batch size.
It is limited by the amount of main memory, which is significantly smaller than disk memory where the entire data set is stored.
However, this issue could be overcome as follows: 
Instead of normalizing each batch with mean and variance computed over the respective batch, we track an exponential average for each domain. 
We compute mean and variance over the batch, update the exponential averages of the respective domain and use these exponential averages for normalization of the batches from the respective domain. 
Over multiple batches, these running statistics will approximate mean and variance as computed over the entire domain and will be representative.
Since this changes the forward pass during training compared to regular batch normalization, it requires working out the derivatives based on the exponential averages of the mean and variance for the backward pass. 
} 

\add{Last but not least, note that Equation \ref{eq:exp_var} computes an unbiased variance estimate.
This means that there is a correction for the bias introduced by estimating a population variance from a finite sample. 
Since the statistic is simply computed from a batch, a correction term is known. 
On the other hand, Equation \ref{eq:exp_inc_var} does not correct for bias. 
We are unaware of a bias correction term in the literature for the \textit{incremental exponentially weighted variance}. 
Yet we do not expect this to influence our method since the values are only significantly different for small sample sizes.  
Sometimes other work also applies a biased variance computed over a sample. 
For instance, regular batch normalization does not correct for bias in the training phase, on purpose, to facilitate gradient computation \cite{ioffe2015batch}.} 

\section{Conclusion} \label{sec:conclusion}
In this work we have presented the first fully unsupervised online personalization approach based on theoretically grounded domain adaptation for accelerometer-based HAR.
The approach incrementally personalizes a general model in real-time, right before classification. 
It also allows to regulate how gradual adaption to new information should be. 

Personalization of general activity recognition models is necessary to achieve good detection results for new unseen users with unique motion patterns.
In the online setting, no samples from the target user are available in advance, but they arrive sequentially, possibly until infinity. 
Thus, an algorithm cannot store all previous observations and retrain the model each time with the entire batch. 
Further, the user’s motion pattern or the operation setting of the system may change over time. 
Therefore, adapting to new information and forgetting old one must be balanced. 
Finally, the target user should not have to do any work to use the recognition system by, say, labeling any activities. 
As we have seen, our approach addresses all of these challenges. 

The experiments on the publicly available WISDM dataset confirmed this. 
Our approach improved accuracy for all but a few users and in particular for users whose movement patterns are quite different from their peers by up to 14 \%. 
This indicates that our approach provides improvements especially for users who are hard to classify by a general model. 
The experiments also showed that utilizing new data as soon as it becomes available is indeed beneficial.
However, depending on the setting, the adaptation rate (momentum) to new information must be stronger or weaker, and may also change over time. 
Investigating an algorithm to automatically regulate the momentum parameter may be a promising future direction.
Last but not least, our experiments showed that using DA-BN layers also leads to competitive results in the supervised and unsupervised batch cases.
This is especially true, if only very little (labeled or unlabeled) target data is available.
A major next step would be to extend these experiments to a variety of additional datasets. 

\section{Acknowledgements}
This work was supported by The International Center for Advanced Communication Technologies (InterACT) and the Baden-Württemberg Foundation. 
This work further used the Extreme Science and Engineering Discovery Environment (XSEDE), which is supported by National Science Foundation grant number ACI-1548562. Specifically, it used the Bridges system, which is supported by NSF award number ACI-1445606, at the Pittsburgh Supercomputing Center (PSC).  

\bibliographystyle{unsrt}  
\bibliography{ms}  

\end{document}